\documentclass[11pt]{article}
\usepackage{coling2020}
\usepackage{times}
\usepackage{url}
\usepackage{latexsym}

\usepackage{amsmath}
\usepackage{amsfonts}
\usepackage{xcolor}
\usepackage{graphicx}
\usepackage{subcaption}
\usepackage{multirow}
\usepackage{booktabs,arydshln}
\usepackage{float}
\usepackage{wrapfig}
\usepackage[english]{babel}
\usepackage{blindtext}
\usepackage{amsfonts, amssymb, amsmath, amsthm}
\usepackage{mathtools}
\usepackage[sans]{dsfont}
\usepackage[ruled,vlined]{algorithm2e}
\newcommand{\Var}{\mathds{V}}

\renewcommand{\epsilon}{\varepsilon}

\newcommand{\bb}{\mathbf{b}}

\newcommand{\defeq}{\vcentcolon=}

\newcommand{\singd}{\text{Single-D}}
\newcommand{\duald}{\text{Dual-D}}

\colingfinalcopy 

\title{Evaluating a Generative Adversarial Framework for Information Retrieval}

\author{Ameet Deshpande \\
  Department of Computer Science \\
  Princeton University \\
  {\tt asd@cs.princeton.edu} \\\And
  Mitesh M. Khapra \\
  Robert-Bosch Centre for\\ Data Science and Artificial Intelligence \\
  Indian Institute of Technology, Madras \\
  {\tt miteshk@cse.iitm.ac.in} \\}

\date{}

\begin{document}
\maketitle
\begin{abstract}
Recent advances in Generative Adversarial Networks (GANs) have resulted in its widespread applications to multiple domains. A recent model, IRGAN, applies this framework to Information Retrieval (IR) and has gained significant attention over the last few years. In this focused work, we critically analyze multiple components of IRGAN, while providing experimental and theoretical evidence of some of its shortcomings. Specifically, we identify issues with the constant baseline term in the policy gradients optimization and show that the generator harms IRGAN's performance. Motivated by our findings, we propose two models influenced by self-contrastive estimation and co-training which outperform IRGAN on two out of the three tasks considered.

\end{abstract}

\section{Introduction}
\label{intro}

%
%

Information Retrieval (IR) can be viewed as a framework which returns a ranked list of documents ($\{d_1,d_2,\dots,d_k\}$) in answer to a query ($q$). This ranked list also implicitly defines a conditional probability distribution for each query ($\mathbb{P}(d|q)$) and captures the intuition that higher-ranked documents should be retrieved more often. This general formulation can be extended to various tasks like web search, content-recommendation, and closed-domain Question-Answering (QA) where information needs, users, and questions are the queries, and web pages, content, and answers are the documents respectively.

At the core, IR induces a probability distribution over documents, and GANs~\cite{goodfellow2014generative} serve as a promising alternative to traditional methods. The generator in a GAN setup is capable of modeling the true probability distribution in high dimensional settings and can be used to retrieve relevant documents for the queries posed, thus making GANs a natural fit for IR. IRGAN~\cite{wang2017irgan} is a popular model which established the first concrete formulation of GANs for IR.

IRGAN consists of a discriminator and a generator, where the discriminator learns to distinguish between documents retrieved by the true probability distribution and the generator's learned probability distribution, while the generator tries to mimic the true probability distribution. Ideally, equilibrium is achieved when the generator manages to rank the documents according to the true distribution. However, IRGAN's loss curves show that equilibrium is not achieved in two of the three tasks.

\paragraph{Contribution} To evaluate the importance of IRGAN's generator, we propose two models inspired by self-contrastive estimation~\cite{goodfellow2014distinguishability} and co-training~\cite{blum1998combining} which outperform IRGAN on two out of the three tasks. We provide a theoretical explanation for the performance degradation of the generator, and our experiments confirm that it is detrimental to IRGAN's performance, rendering it equivalent to sub-optimal noise-contrastive estimation methods~\cite{gutmann2010noise}. Given the usefulness of GANs, we believe that a critical evaluation of adversarial frameworks for IR is necessary, and we hope that our study provides a foundation for the same.
\section{Related Work} \label{sec:related}

\paragraph{Noise Contrastive Estimation (NCE)}
NCE~\cite{gutmann2010noise} is a parameter estimation method used to train models to differentiate between true data instances and noise samples. NCE can be shown to be asymptotically unbiased~\cite{dyer2014notes} and provides an alternative way to approximate traditional maximum-likelihood estimation (MLE) based retrieval models~\cite{baeza1999modern,zhai2001model,hofmann1999probabilistic}. Self-contrastive estimation~\cite{goodfellow2014distinguishability} uses the same model for learning and generating the noise distribution, and dual-learning~\cite{he2016dual} can be perceived as a co-operative setup where one model generates the noise distribution for the other.

\paragraph{Generative Adversarial Networks}
GANs \cite{goodfellow2014generative} are generative models \cite{salakhutdinov2010efficient} which avoid computing intractable normalization constants in probability distributions. The \textit{generator} tries to implicitly model the true data distribution and the \textit{discriminator} learns to differentiate between true and generated data points. GANs have been widely applied to various problems like image generation~\cite{radford2015unsupervised,zhu2017unpaired,ledig2017photo}, text generation~\cite{seqgan,fedus2018maskgan}, and cross-modal retrieval~\cite{peng2019cm}.

\paragraph{Adversarial frameworks for IR} IRGAN~\cite{wang2017irgan} uses GANs to learn models for web search, recommendation, and QA. \cite{he2018adversarial} introduce adversarial perturbations for robust ranking for recommendation.~\cite{yang2019adversarial} modify IRGANs for QA and~\cite{park2019adversarial} use a semi-supervised approach to generate adversarial samples which make the model robust and sample efficient. Since IRGAN is a widely adopted model, it forms the basis of our analysis.





\section{Background} \label{sec:background}
We present important details of IRGAN in this section, and refer the reader to~\cite{wang2017irgan} for a more detailed explanation. In the subsequent sections, $D$ denotes the discriminator, $G$ the generator, $p_{true}$ the real probability distribution over documents, $\phi$ the parameters of $D$, $\theta$ the parameters of the $G$, $d$ the document, $q$ the query and $r$ the rank of $d$ with respect to a $q$. $f$ is $D$'s model, and $D(d|q_n)=\sigma(f_{\phi}(d,q))$.
\paragraph{Minimax objective} Just like in GANs, IRGAN-Pointwise uses a joint objective.

\vspace{-1.5em}\begin{align}\label{eq:irgan} J^{G^*,D^*} &= \min_{\theta} \max_{\phi} \sum_{n=1}^{N} \left (\mathbb{E}_{d\sim p_{true}(d|q_n,r)}[ \log D(d|q_n)] + \mathbb{E}_{d\sim p_{\theta}(d|q_n,r)}[ \log \left ( 1 - D(d|q_n)\right)]\right ) \end{align}\vspace{-1em}

The first term increases the likelihood of samples from $p_{true}$ and the second decreases it for $G$'s learned distribution $p_{\theta}(d|q_n,r)$. IRGAN-Pairwise is an alternate formulation where the $(d,q)$ pairs are substituted by triples $(d_i,d_j,q)$ in equation~\ref{eq:irgan}, where document $d_i$ is more relevant to $q$ than $d_j$.

\paragraph{Optimization} The discriminator essentially performs binary classification and can be optimized using backpropagation~\cite{rumelhart1986learning}. However, unlike in the original GAN setup~\cite{goodfellow2014generative}, the generator in IRGAN has a discrete sampling step, because it samples a document from an accessible pool. Thus, REINFORCE~\cite{williams1992simple}, a policy gradients approach, is used to calculate the gradients for the generator. The update averaged over $K$ documents is given below.

\vspace{-1.5em}\begin{align} \label{eq:reinforce}  &\nabla_{\theta}J^{G}(q_n) =\frac{1}{K} \sum_{k=1}^K \nabla_{\theta} \log p_{\theta}(d_k|q_n,r) \log\left (1+exp\left(f_{\phi}\left(d_k,q_n\right)\right)\right)\end{align}\vspace{-1em}

REINFORCE's gradient updates generally have high variance, and reward baselines are used to make the learning stable~\cite{weaver}. A common baseline is the value function of the state (here, query) $\equiv \mathbb{E}_{d\sim p_{\theta}(d|q_n,r)}[\log(1+exp(f_{\phi}(d_k,q_n)))]$. Since this expectation over all the documents is intractable to compute, IRGAN uses a constant baseline of $0.5$ for all the queries (appendix B~\cite{wang2017irgan}), and to alleviate training issues, $\log(1+exp(f_{\phi}(d_k,q_n)))$ is replaced with $\sigma(f_{\phi}(d_k,q_n))$. The final gradient update is as follows.

\vspace{-1.5em}\begin{align} \label{eq:final}  &\nabla_{\theta}J^{G}(q_n) =\frac{1}{K} \sum_{k=1}^K \nabla_{\theta} \log p_{\theta}(d_k|q_n,r)\times 2\left (\sigma(f_{\phi}(d_k,q_n))-0.5 \right )\end{align}\vspace{-1em}

\section{Tasks and Evaluation}

We follow IRGAN and evaluate on Web Search, Item Recommendation, and Question Answering (QA). The datasets used are LETOR~\cite{letor}, Movielens~\cite{harper2015movielens} and InsuranceQA~\cite{feng2015applying} respectively (appendix~\ref{app:dataset}). We report the NDCG@5 and Precision@5 metrics for Web Search and Item Recommendation, and the Precision@1 metric for QA.

\section{Models and Method} \label{sec:models}
\begin{figure}[h]
\begin{subfigure}{.45\textwidth}
  \centering
  \includegraphics[width=\linewidth]{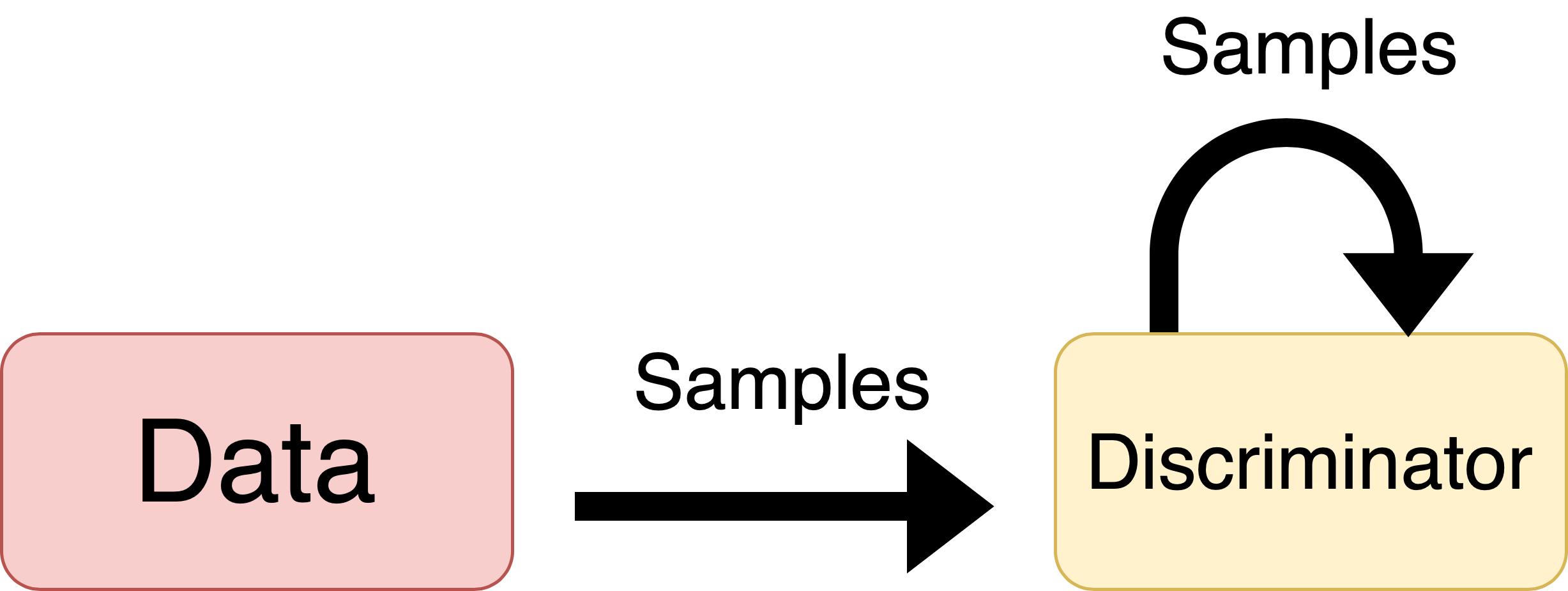}
\end{subfigure}\hfill
\begin{subfigure}{.45\textwidth}
  \centering
  \includegraphics[width=\linewidth]{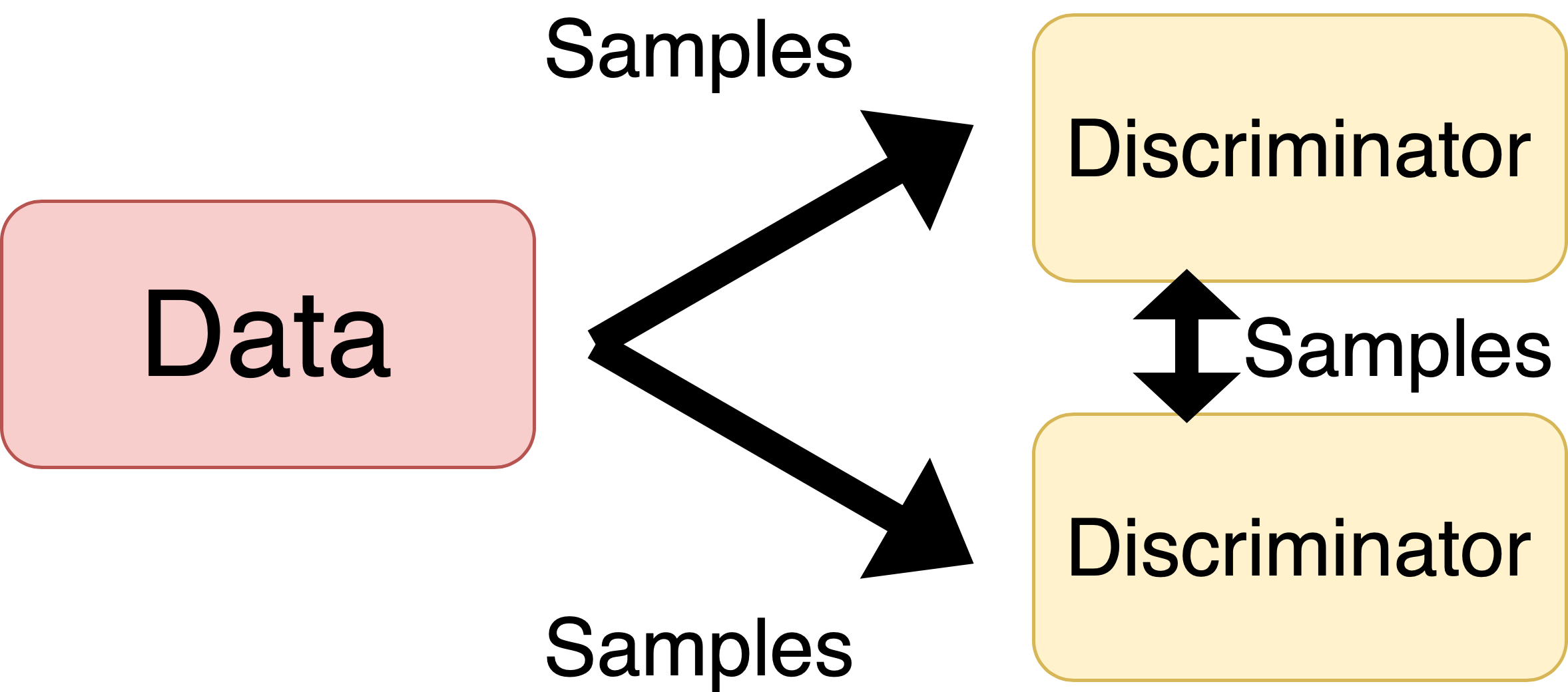}
\end{subfigure}
\caption{Proposed Models \singd{} (left) and \duald{} (right). \textit{\singd{}} uses a single discriminator which feeds itself negative samples, and \textit{\duald{}} feeds negative samples to each other}
\label{fig:models}
\end{figure}

IRGAN's setup can be considered as a dynamic negative sampling~\cite{zhang2013optimizing} approach where the generator continuously adapts the negative samples that it feeds to the discriminator. However, these negative samples can come from other sources, and we propose two different models based on the same. The first is the \textit{Single Discriminator (\singd{})} model motivated by self-contrastive estimation~\cite{goodfellow2014distinguishability} where negative samples come from the model's ($M$) probability distribution. $M$, like $D$ is a discriminator, and the probability of sampling a document $d_i\in \mathcal{D}$ according to its distribution is $\dfrac{M(d_i|q_n)}{\sum_{d\in \mathcal{D}} M(d|q_n)}$. The second is a two model setup called \textit{Dual Discriminator (\duald{})} and is motivated by co-training~\cite{blum1998combining}. It is similar to \textit{\singd{}}, but instead of the models feeding negative samples to themselves, they feed them to each other. One of the models is randomly chosen at evaluation time. The positive samples are drawn from the true data distribution for both \singd{} and \duald{}. Figure~\ref{fig:models} illustrates both the proposed models.

We follow IRGAN and compare RankNet~\cite{burges2005learning}, LambdaRank~\cite{burges2007learning}, IRGAN-pointwise, and IRGAN-pairwise on web search, BPR~\cite{goodfellow2014generative}, LambdaFM~\cite{yuan2016lambdafm}, and IRGAN-pointwise on item recommendation, and QA-CNN~\cite{santos2016attentive}, LambdaCNN~\cite{zhang2013optimizing,santos2016attentive}, and IRGAN-pairwise on QA. Our models \textit{\singd{}} and \textit{\duald{}} are evaluated on all the tasks considered. We report the best performing hyperparameters in appendix~\ref{app:hyperparameters}.

\section{Experiments and Discussion}\label{sec:experiments}

\paragraph{Results}
Table~\ref{table:results} summarizes all our experiments. On \textbf{web search}, \textit{\duald{}} outperforms both the variants of IRGAN, while \textit{\singd{}} matches the performance of the better variant. The same applies to the \textbf{QA} task where \textit{\duald{}} performs slightly better than IRGAN-pairwise while \textit{\singd{}} matches its performance\footnote{The numbers differ slightly from IRGAN~\cite{wang2017irgan}. After close correspondence with its authors, we obtained all the random seeds used by the models, but the results for QA could not be reproduced. We mention the results on our random seeds, and fully believe that any random seed which gives better performance for IRGAN should do so for our model as well.}. The strong performance of \textit{\singd{}}, which unlike IRGAN contains a single model, shows that the generator in IRGAN might not be important for its performance improvements. \textbf{Item-recommendation} is the only task where IRGAN performs better than \textit{\singd{}} and \textit{\duald{}}. However, the performance difference between IRGAN and \textit{\singd{}} is negligible and corresponds to it making just 7 more mistakes on a test set of 943 users. We believe that \textit{\duald{}} performs better than \textit{\singd{}} on two variants for the same reason that ensembles perform better than single classifiers~\cite{dietterich2000ensemble,dvzeroski2004combining}. One model helps correct the errors being made by the other model to some extent.

\begin{table*}[t]
\centering
\resizebox{\linewidth}{!}{
\begin{tabular}{cccccccc}\toprule

\multicolumn{3}{c}{\textbf{Web Search}} & \multicolumn{3}{c}{\textbf{Recommendation}} & \multicolumn{2}{c}{\textbf{Question Answering}} \\ \cmidrule(lr){1-3}  \cmidrule(lr){4-6}  \cmidrule(lr){7-8}

\textbf{Model} & \textbf{P@5} & \textbf{NDCG@5} & \textbf{Model} & \textbf{P@5} & \textbf{NDCG@5} & \textbf{Model} & \textbf{P@1} \\ \cmidrule(lr){1-3}  \cmidrule(lr){4-6}  \cmidrule(lr){7-8}
\textbf{RankNet} & 0.1219 & 0.1709 & \textbf{BPR} & 0.3044 & 0.3245 & \textbf{QA-CNN} & 0.613 \\
\textbf{LambdaRank} & 0.1352 & 0.1920 & \textbf{LambdaFM} & 0.3474 & 0.3749 & \textbf{LambdaCNN} & 0.629 \\ \midrule
\textbf{IRGAN-pointwise} & 0.1657 & 0.2225 &  & \textbf{0.3750} & \textbf{0.4099} &  & - \\
\textbf{IRGAN-pairwise} & 0.1676 & 0.2154 &  & - & - & &0.616 \\
\textbf{\singd{}} & 0.1676 & 0.2190 & & 0.3675 & 0.3925 & & 0.614 \\
\textbf{\duald{}} & \textbf{0.1733} & \textbf{0.2252} & & 0.3450 & 0.3730 & & \textbf{0.623} \\

\bottomrule
\end{tabular}
}
\caption{\duald{} outperforms all IRGAN variants on web search and QA, and \singd{} matches the performance on all tasks. Its worse performance on recommendation corresponds only to $7$ more errors}
\vspace{-1em}
\label{table:results}
\end{table*}

\begin{wrapfigure}{r}{0.5\textwidth}
  \vspace{-1.5em}
  \centering
  \strut
    \includegraphics[width=0.5\textwidth]{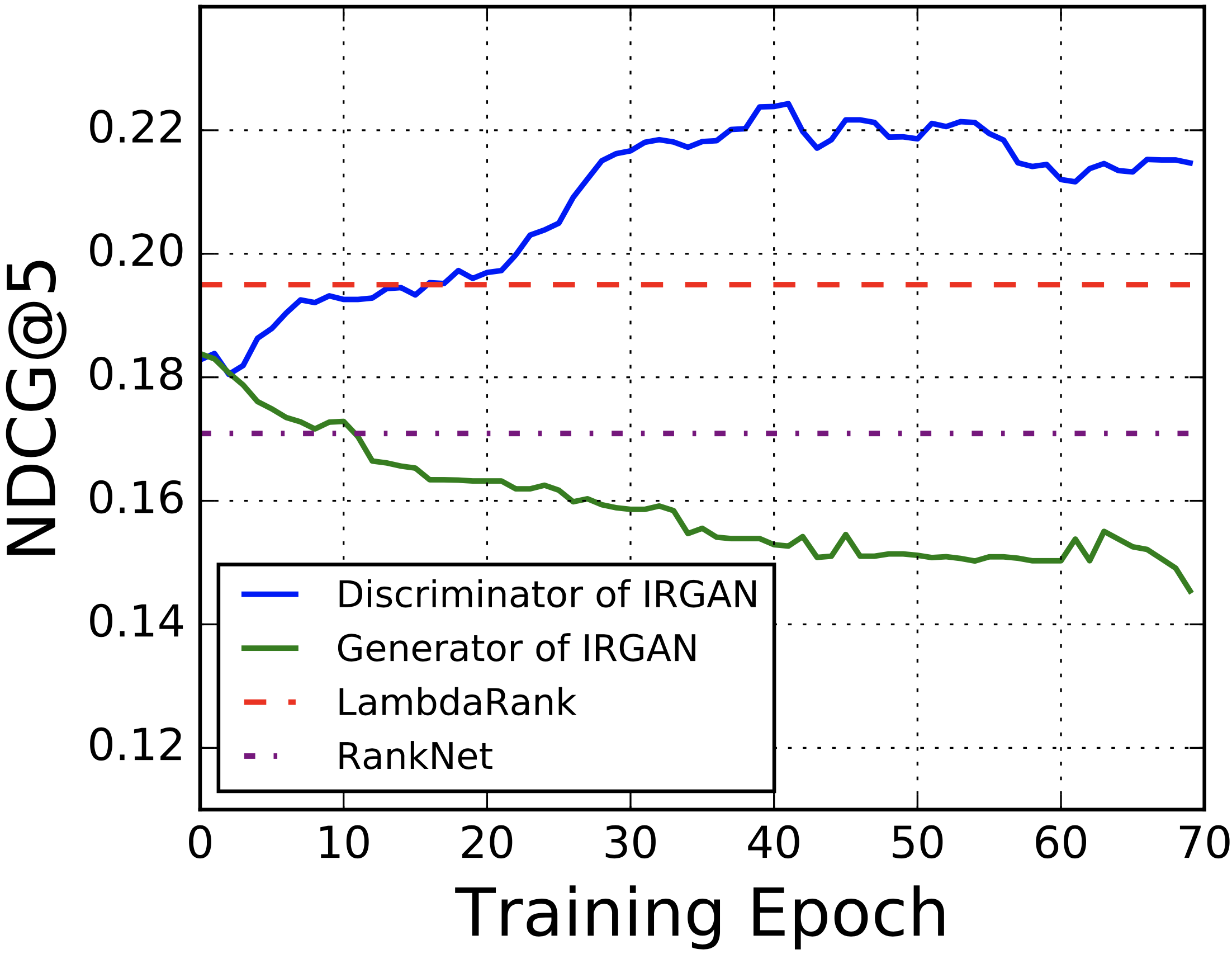}
  \vspace{-2em}
	\caption{Loss curves for IRGAN on web search task. The generator is initialized with a pre-trained model, and its performance degrades throughout training, which is contrary to expectation.}
  \label{fig:ndcg}
\end{wrapfigure}

\paragraph{Loss Curves} The loss curves for IRGAN reported in~\cite{wang2017irgan} follow a peculiar trend in both web search and QA (figure~\ref{fig:ndcg} illustrates an example). The generator is initialized with a pre-trained model, and its performance degrades throughout training, which is contrary to what one would expect in GANs, where the generator's performance should improve till equilibrium. At that stage, the discriminator cannot differentiate between the true data distribution and generator's learned distribution~\cite{goodfellow2014generative}. Negative samples generated from a degrading generator continuously deviate from the true distribution it needs to learn as training progresses, and this makes the discriminator's objective easier, thus hurting IRGAN's performance. This is equivalent to a sub-optimal NCE setup where the quality of negative samples is bad. However, in both \textit{\singd{}} and \textit{\duald{}}, the performance of the models improves throughout training, thus improving the quality of negative samples.


\paragraph{Baseline} As shown in equation~\ref{eq:final}, IRGAN uses a constant baseline of $0.5$ as an approximation of the value function. We show that this may not be the best choice, and our observations are in line with~\cite{greensmith2004variance}. Our theoretical result in appendix~\ref{app:proof} makes plausible assumptions and shows that using a constant baseline increases the variance of gradient updates for tasks where the fraction of correct documents that can be retrieved per query is low (equation~\ref{eq:variance}). Since the average number of correct documents per query for QA ($0.002$) and web search ($0.005$) is significantly lower than for item-recommendation ($0.015$), the constant baseline affects the generator's training in both web search and QA by increasing the variance and as a result making convergence harder. This is empirically supported by better performance of \duald{} and equal performance of \singd{} on both those tasks, even though they don't use a generator, and only discriminators. The loss curves in IRGAN (Figures 3,8 in~\cite{wang2017irgan}) which show the deteriorating generator further corroborate our result that the constant baseline term harms the generator.

\vspace{-1em}\begin{align}\label{eq:variance}
&\Var(g(b))\geq\nonumber\\
&b^2\left(\frac{Q_{max}}{b}-1\right)^2\mathbb{E}_{\rho_\pi}\left[\mathbb{P}(a\in\mathcal{A}_1(s_t))\mathbb{E}_{\pi{(\mathcal{A}_1})}[||\nabla_\theta \log \pi_\theta(a_t|s_t)-\mathbb{E}_{\rho_\pi,\pi}\left[\nabla_\theta \log \pi_\theta(a_t|s_t)\right]||^2]\right]
\end{align}


\section{Conclusion}\label{sec:conclusion}
In this work, we theoretically and experimentally show issues with the adversarial framework of a popular IR model. Through experiments using our proposed models which outperform IRGAN on two out of three tasks and our theoretical analysis of the variance in the policy gradients update, we show that the generator in IRGAN is harmful for its learning, thus converting IRGAN into a sub-optimal NCE model. Strong results of IRGAN on the recommendation task shows that adversarial learning is a promising area for IR when applied carefully, and we hope that our study provides a solid foundation for fundamental research in this area.

\bibliographystyle{coling}
\bibliography{coling2020}

\appendix
\section{Variance of Gradient Updates in REINFORCE}\label{app:proof}

We proceed to lower bound the variance of REINFORCE's updates while making plausible assumptions. We use standard notation where $S$ is the state space, $s_t$ is the state at time $t$, $\mathcal{A}$ is the action space, $\pi$ is the policy, $\theta$ represents the parameters of the model, $\hat{Q}$ is the $Q$-value, and $\rho_\pi$ is the state-visitation frequency. Also, in 

Let $\bb$ be a vector of size $|S|$ which denotes the baseline term used in REINFORCE's updates and $S$ be the state space. Let $\bb(s_t)$ represent the baseline for value for the state $s_t$. $g(\bb)$ describes the gradient update~\cite{wu2018variance}, and is a function of the baseline.
\begin{align*}
g(\bb)\defeq \nabla_\theta \log \pi_\theta(a_t|s_t)\left(\hat{Q}(s_t,a_t)-\bb(s_t)\right), \qquad a_t\sim\pi_\theta(\cdot|s_t),\qquad s_t\sim\rho_\pi(\cdot)
\end{align*}
Assume that the baseline term is the same for all the states, which is indeed the case for IRGAN.
\begin{align*}
\bb(s_i)=b \qquad \forall s_i\in S
\end{align*}
We can rewrite the function $g(\cdot)$ as
\begin{align*}
g(b)\defeq \nabla_\theta \log \pi_\theta(a_t|s_t)\left(\hat{Q}(s_t,a_t)-b\right)
\end{align*}
The variance of the gradient $g(\cdot)$ is given by
\begin{align*}
\Var(g(b))=\mathbb{E}_{\rho_\pi,\pi}\left[||g(b)-\mathbb{E}_{\rho_\pi,\pi}\left[g(b)\right]||^2\right]
\end{align*}
Let $\mathcal{A}$ be the set of actions available in each state. For each state $s\in S$, $\mathcal{A}$ can be partitioned into $\mathcal{A}_1(s)$ and $\mathcal{A}_2(s)$ which are actions partitions such that $Q$ value of picking that action is less than the baseline and greater than the baseline respectively.
\begin{align*}
\forall a\in \mathcal{A}_1(s): \hat{Q}(s,a)<b\\
\forall a\in \mathcal{A}_2(s): \hat{Q}(s,a)\geq b
\end{align*}
The variance of $g(\cdot)$ can then be simplified to the following, where $\pi{(\mathcal{A}_1})$ refers to the policy which picks actions only from the set $\mathcal{A}_1(s)$ when in state $s$.
\begin{align*}
\Var(g(b))=\mathbb{E}_{\rho_\pi}\left[\mathbb{P}(a\in\mathcal{A}_1(s_t))\mathbb{E}_{\pi{(\mathcal{A}_1})}[||g(b)-\mathbb{E}_{\rho_\pi,\pi}\left[g(b)\right]||^2]\right]+\\
\mathbb{E}_{\rho_\pi}\left[\mathbb{P}(a\in\mathcal{A}_2(s_t))\mathbb{E}_{\pi_{(\mathcal{A}_2)}}[||g(b)-\mathbb{E}_{\rho_\pi,\pi}\left[g(b)\right]||^2]\right]
\end{align*}
At the beginning of training, we can make the following assumption
\begin{align*}
\mathbb{P}(a\in\mathcal{A}_1(s_t))\gg\mathbb{P}(a\in\mathcal{A}_2(s_t))
\end{align*}
This is because the number of ``correct" documents corresponding to a given query is very low, and the policy at the beginning of training is uniformly random or bad. This makes the probability of picking the good actions (correct documents) low. This reduces the variance expression to 
\begin{align*}
\Var(g(b))\approx\mathbb{E}_{\rho_\pi}\left[\mathbb{P}(a\in\mathcal{A}_1(s_t))\mathbb{E}_{\pi{(\mathcal{A}_1})}[||g(b)-\mathbb{E}_{\rho_\pi,\pi}\left[g(b)\right]||^2]\right]
\end{align*}
Define $Q_{max}$ as 
\begin{align*}
Q_{max}=\max_{s_t\in S}\max_{a_t\in \mathcal{A}_1(s_t)}\hat{Q}(s_t,a_t)
\end{align*}
Then, by pulling out the factor $\left(Q_{max}-b \right)^2$ which is a constant, we have
\begin{align*}
\Var(g(b))\geq \left(Q_{max}-b \right)^2\mathbb{E}_{\rho_\pi}\left[\mathbb{P}(a\in\mathcal{A}_1(s_t))\mathbb{E}_{\pi{(\mathcal{A}_1})}[||\nabla_\theta \log \pi_\theta(a_t|s_t)-\mathbb{E}_{\rho_\pi,\pi}\left[\nabla_\theta \log \pi_\theta(a_t|s_t)\right]||^2]\right]
\end{align*}
We now have that (1) the term in expectation is independent of $b$, (2) $b>Q_{max}$ $\forall a\in \mathcal{A}_1$, and (3) term in the expectation is always positive. We have the following, where $lower(\Var(g(b)))$ denotes the lower bound.
\begin{align*}
\quad b_1>b_2 \implies \left(Q_{max}-b_1 \right)^2>\left(Q_{max}-b_2 \right)^2 \implies lower(\Var(b_1))>lower(\Var(b_2)) \quad \text{given}\quad Q_{max}
\end{align*}
The lower bound on variance can be rewritten as follows
\begin{align*}
&\Var(g(b))\geq\\
&b^2\left(\frac{Q_{max}}{b}-1\right)^2\mathbb{E}_{\rho_\pi}\left[\mathbb{P}(a\in\mathcal{A}_1(s_t))\mathbb{E}_{\pi{(\mathcal{A}_1})}[||\nabla_\theta \log \pi_\theta(a_t|s_t)-\mathbb{E}_{\rho_\pi,\pi}\left[\nabla_\theta \log \pi_\theta(a_t|s_t)\right]||^2]\right]
\end{align*}
If for two tasks, $Q_{{max}_1}<Q_{{max}_2}$ implies $\Var(g(b)|Q_{{max}_1})>\Var(g(b)|Q_{{max}_2})$.
Since $Q_{max}$ is typically lower for tasks where the fraction of correct documents that can be retrieved are low, we have that the lower bound is higher for such tasks. This is equivalent to low $Q$-values of bad actions in large action spaces with sparse rewards~\cite{andrychowicz2017hindsight}. Intuitively, when only a small fraction of documents are correct, it is harder for the generator to fool the discriminator, and hence the $Q$ values corresponding to incorrect documents are low. While we have proved a lower bound result, it provides some intuition as to why the same baseline term might have different effects on different tasks.

\section{Dataset Statistics}\label{app:dataset}

\begin{table}[h]
\begin{center}
\begin{tabular}{ ccc }
    \toprule
    \textbf{Task} & \textbf{Dataset} & \textbf{Number of queries}\\
    \midrule
    Web Search & LETOR \cite{liu2007letor} & 784\\
    Recommendation & Movielens~\cite{harper2015movielens} & 943\\
    Question Answering & InsuranceQA \cite{feng2015applying} & 12887\\
    \bottomrule
\end{tabular}
\end{center}
\label{tab:datasets}
\caption{Datasets}
\end{table}

\section{Hyperparameters} \label{app:hyperparameters}

\begin{table}[h]
\begin{center}
\begin{tabular}{ cc }
    \toprule
    \textbf{Hyperparameter} & \textbf{Description}\\
    \midrule
    Learning Rate & For both generator and discriminator\\
    Batch Size & Batch size for training\\
    Embed Dim & Embedding dimension of words\\
    Outer Epochs & Number of epochs of training\\
    Inner Epochs & The number of epochs \duald{} models\\
    &are trained for each outer epoch\\
    Temperature & Temperature parameter for softmax\\
    Random Seed & The random seed used for initializations \\
    Feature Size & The intermediate size of neural networks \\
    DNS\_K & The number of negative samples \\
    \bottomrule
\end{tabular}
\end{center}
\label{tab:hyperparameters}
\caption{Hyperparameters for our models}
\end{table}

\begin{table}[h]
\begin{center}
\begin{tabular}{ ccc }
    \toprule
    \textbf{Hyperparameter/Seed} & \textbf{Range/List} & \textbf{Best}\\\midrule
    Learning Rate & 0.002-0.2 & 0.004\\
    Batch Size & [8,16,32] & 8\\
    Feature Size & [46, 92] & 46\\
    Random Seed & [20,40,60] & 40\\
    \bottomrule
\end{tabular}
\end{center}
\label{tab:web_single}
\caption{\singd{} for web search}
\end{table}

\begin{table}[h]
\begin{center}
\begin{tabular}{ ccc }
    \toprule
    \textbf{Hyperparameter/Seed} & \textbf{Range/List} & \textbf{Best}\\\midrule
    Learning Rate & 0.002-0.2 & 0.006\\
    Outer Epochs & [30,50] & 50\\
    Inner Epochs & [30,50] & 30\\
    Batch Size & [8,16,32] & 8\\
    Feature Size & [46, 92] & 46\\
    Random Seed & [20,40,60] & 40\\
    \bottomrule
\end{tabular}
\end{center}
\label{tab:web_co}
\caption{\duald{} for web search}
\end{table}

\begin{table}[h]
\begin{center}
\begin{tabular}{ ccc }
    \toprule
    \textbf{Hyperparameter/Seed} & \textbf{Range/List} & \textbf{Best}\\\midrule
    Learning Rate & 0.01-0.05 & 0.02\\
    Batch Size & 10 & 10\\
    Embedding Dimension & [20, 40, 60] & 20\\
    Random Seed & 70 & 70\\
    DNS\_K & 5  & 5\\
    \bottomrule
\end{tabular}
\end{center}
\label{tab:content_single}
\caption{\singd{} for item recommendation}
\end{table}

\begin{table}[h]
\begin{center}
\begin{tabular}{ ccc }
    \toprule
    \textbf{Hyperparameter} & \textbf{Best}\\\midrule
    Learning Rate & 0.05\\
    Epochs & 20 \\
    Batch Size & 100\\
    Embedding Dimension & 100\\
    \bottomrule
\end{tabular}
\end{center}
\label{tab:qa_single}
\caption{\singd{} for Question Answering}
\end{table}

\begin{table}[h]
\begin{center}
\begin{tabular}{ ccc }
    \toprule
    \textbf{Hyperparameter} & \textbf{Best}\\\midrule
    Learning Rate & 0.05\\
    Outer Epochs & 20\\
    Inner Epochs & 1 \\
    Batch Size & 100\\
    Embedding Dimension & 100\\
    \bottomrule
\end{tabular}
\end{center}
\label{tab:qa_co}
\caption{\duald{} for Question Answering}
\end{table}


\end{document}